\documentclass[a4paper,twoside]{article}

\usepackage{float}
\usepackage{stfloats}
\usepackage{epsfig}
\usepackage{subcaption}
\usepackage{calc}
\usepackage{amstext}
\usepackage{amsthm}
\usepackage{multicol}
\usepackage{pslatex}
\usepackage{comment}
\usepackage{apalike}
\usepackage[linesnumbered,ruled,vlined]{algorithm2e}
\SetKwInput{KwInput}{Input}
\SetKwInput{KwOutput}{Output}
\usepackage[bottom]{footmisc}
\usepackage{comment}
\usepackage{hyperref}
\usepackage[export]{adjustbox}
\usepackage{tikz}
\usetikzlibrary{shapes.geometric, arrows}
\usepackage{todonotes}

\usepackage{caption}
\usepackage[none]{hyphenat}
\usepackage{doi}

\usepackage{amsmath,amssymb,amsfonts}
\usepackage{algorithmic}
\usepackage{graphicx}
\usepackage{textcomp}
\usepackage{xcolor}
\usepackage{listings}
\usepackage{SCITEPRESS}     % Please add other packages that you may need BEFORE the SCITEPRESS.sty package.

\begin{document}

% other possible title(s) [please add]:
%Emergent Dynamics of Electric Vehicle Charging: An Agent-Based Simulation of Scheduling, Infrastructure Mix, and Grid Constraints
\title{A Grid-Aware Agent-Based Model for Analyzing Electric Vehicle Charging Systems}
\author{\authorname{Khalil Al-Rahman Youssefi\sup{1,2}\orcidAuthor{https://orcid.org/0000-0002-8719-7699}, Marija Gojković\sup{1,2}\orcidAuthor{https://orcid.org/0009-0002-4618-8929}, Walter Stefanutti \sup{3}\orcidAuthor{https://orcid.org/0009-0004-7393-7424}, Mika Auer\sup{1,2}\orcidAuthor{https://orcid.org/0009-0007-2754-8075} and Melanie Schranz\sup{1}\orcidAuthor{https://orcid.org/0000-0002-0714-6569}}
\affiliation{\sup{1}Lakeside Labs, Lakeside Park B04b, Klagenfurt, Austria}
\affiliation{\sup{2}Alpen-Adria-Universität, Universitätsstraße 65–67, Klagenfurt, Austria}
\affiliation{\sup{3}Silicon Austria Labs, Power Electronics division, High Tech Campus Villach - Europastraße 12, Austria}
\email{\{youssefi, auer, gojkovic, schranz\}@lakeside-labs.com \\ walter.stefanutti@silicon-austria.com}
}


\abstract{
This paper presents a configurable, grid-aware Agent-Based Model (ABM) for the systematic analysis of electric vehicle (EV) charging systems under configurable infrastructure and operational conditions. The model integrates heterogeneous EV behavior, charging column constraints, and a shared Energy Sandbox that regulates aggregate power allocation, enabling the joint study of user-centric charging dynamics and facility-level power behavior. Implemented in Python using the SimPy discrete-event framework, the approach supports scalable, event-driven simulations across varying system sizes, charger compositions, and scheduling strategies. A representative workplace charging scenario is investigated to illustrate how infrastructure configuration and coordination mechanisms influence energy delivery performance, infrastructure utilization, and aggregate load characteristics. The results highlight the context-dependence of infrastructure suitability and demonstrate how charging strategies and charger types reshape both service-level outcomes and grid-facing behavior. The proposed ABM provides a flexible and extensible simulation environment for exploring technical, operational, and grid-aware aspects of EV charging ecosystems, and for serving as a methodological basis for subsequent studies on advanced coordination strategies beyond the specific scenario analyzed in this study.
}
\onecolumn \maketitle \normalsize \setcounter{footnote}{0} \vfill

\section{\uppercase{Introduction}}
\label{sec:intro}

The rapid electrification of mobility is transforming both transportation systems and electrical power networks. While electric vehicles (EVs) are essential for decarbonizing mobility, their large-scale adoption introduces new operational challenges for charging infrastructure and distribution grids. In contrast to conventional loads, EV charging demand is inherently stochastic, driven by human mobility patterns, heterogeneous vehicle characteristics, and temporal constraints such as arrival times and parking durations. As EV penetration increases, unmanaged charging behavior can lead to synchronized high-power demand peaks, congestion at charging facilities, and violations of grid operating limits, up to and including a blackout.
Current public charging solutions remain insufficient both in spatial coverage and in their ability to flexibly respond to dynamic energy conditions. Emerging concepts such as shared charging infrastructures aim to address this gap by enabling charging wherever vehicles are naturally parked and by actively coordinating energy allocation among multiple users. However, designing such systems requires understanding complex interactions between EV users, charging infrastructure, and grid constraints

Agent-Based Modeling (ABM) provides a framework for studying these dynamics: By representing individual EVs as autonomous agents with heterogeneous behaviors and charging needs, ABMs allow system-level phenomena, such as demand synchronization, charging high rates of renewable energy source at low prices, to emerge from decentralized interactions. At the same time, charging columns (CCs) introduce infrastructure-level constraints, including limited ports and heterogeneous power ratings, which strongly influence aggregate load trajectories. Capturing the coupled behavior of these components is essential for evaluating operational strategies under realistic conditions.

In this work, we present an ABM to simulate the dynamics of EV charging systems under grid-aware operation. The contribution of this work is threefold. First, it provides a configurable, grid-aware ABM framework capable of jointly evaluating EV-centric, infrastructure-centric, and grid-centric performance indicators across heterogeneous charging scenarios. Second, it demonstrates, using a representative workplace charging use case, how infrastructure composition, fleet scale, and representative rule-based charging strategies shape emergent charging dynamics and system feasibility. Third, it provides the methodological foundation for subsequent studies on advanced coordination mechanisms, including self-organizing and learning-based charging strategies, within a common and extensible simulation environment.

The paper is organized as follows. Section~\ref{sec:related_work} reviews related work. 
Section~\ref{sec:system_model} presents the proposed system model. 
Section~\ref{sec:system_analysis} provides the use-case definition, experimental setup, and a detailed system-level analysis of the simulation results.
Section~\ref{sec:conclusion} concludes the paper.
\section{\uppercase{Related Work}}
\label{sec:related_work}

This section reviews (i) simulation methodologies used to study shared EV charging and (ii) evaluation metrics and decentralized coordination strategies that shape system behavior, highlighting the gaps that motivate the presented modeling framework.

\subsection*{Simulation Methodologies for Shared EV Charging}

Agent‑based modeling has proven effective for analyzing complex systems in which heterogeneous actors and local decision rules generate emergent, system‑level behavior. Its use across domains such as semiconductor manufacturing~\cite{enhanced2025youssefi}, production logistics~\cite{gojkovic2026preserving} or edge‑computing systems~\cite{Schranz2024ABMEdgeContinuum} demonstrates its strength in capturing decentralized dynamics and stochastic interactions. Such capabilities are increasingly relevant for understanding shared EV charging environments as well.

Notably, comparisons of event-driven and time-stepped simulation show that fixed time increments can introduce substantial numerical distortions unless chosen extremely small, whereas discrete-event approaches avoid these issues and yield higher accuracy and efficiency~\cite{Buss2010Comparison,Pedrielli2012DESvsTimeStep,SoftwareSim2022DTvsDES}.

Building on these methodological foundations, research on grid‑coupled EV charging has developed simulation platforms that integrate charging behavior with detailed distribution‑system models. Grid‑aware co‑simulation environments such as EV‑EcoSim~\cite{Balogun2024EVEcoSim} couple EV charging with storage, photovoltaic generation, transformer dynamics, and power‑flow solvers to assess system‑level impacts under different control strategies. Along similar lines, unified OpenDSS‑based~\cite{Iranpour2025UnifiedCoSim} framework constructs feeder‑level digital twins from synthetic bottom‑up data, enabling the analysis of thermal loading risks and voltage deviations as EV penetration continues to rise. Complementing these grid-centric tools, coupled traffic–power simulations demonstrate how clustered arrivals and realistic mobility patterns can drive localized congestion and reverse-power-flow issues in suburban feeders~\cite{Obinata2024USIM}. At the same time, lightweight open-source simulators and load–profile generators enable rapid prototyping of tariff and policy scenarios, often by imposing aggregate grid constraints rather than resolving detailed electrical dynamics~\cite{EVLPG2019GitHub}.

\subsection*{Evaluation Metrics, Load Quality, and Decentralized Coordination}

The literature typically distinguishes between grid‑centric and user‑centric performance indicators. Grid‑centric measures such as peak demand, transformer loading, and basic thermal or voltage‑limit checks remain standard for assessing operational feasibility~\cite{Maghami2025DiscoverSustainability}. Recent work focuses on stability‑oriented metrics, particularly load‑variance and ramp‑rate constraints, which capture short‑term fluctuations linked to equipment stress and system volatility~\cite{Xu2025IJLCTCapacityPlanning}. On the user side, state‑of‑charge (SoC) indicators such as final SoC or unmet energy provide a minimal yet informative basis for assessing service quality~\cite{Li2023MARLCharging}. However, these metrics alone do not reflect grid stability, motivating analyses that link SoC outcomes with aggregate power‑fluctuation behavior.

A growing body of research emphasizes the quality and predictability of aggregate EV demand. Ramp constraints mitigate short-term volatility and protect distribution assets~\cite{Xu2025IJLCTCapacityPlanning}, while predictability is increasingly crucial for day‑ahead and intra‑day operations, improving forecast accuracy and reducing procurement risk~\cite{Ostermann2024ProbEVForecast,MuratoriYip2024NREL89775}. Engineered EV load shapes aligned with renewable generation have also been proposed to improve grid integration~\cite{EPRI2020EVLoadShapes,HuamanRivera2024ElectronicsLoadModeling}.

Within decentralized coordination, agent-based and multi-agent approaches model heterogeneous user behavior, local information, and asynchronous decisions. Traffic–power coupled studies show that clustered charging events can create spatial bottlenecks even at moderate EV adoption~\cite{Plagowski2021AgentGridTraffic}. Multi-agent simulations of tariff-following and real-time pricing reveal earlier overload onset under naive price responsiveness, underscoring the need for coordination mechanisms balancing user cost and grid headroom~\cite{Christensen2024MABSEV}. High‑resolution residential models further show that second‑level clustering under real-time pricing can accelerate transformer degradation, while coordinated EV–PV–battery portfolios delay overload incidence~\cite{Cong2025MABS}.

Despite substantial progress, several gaps persist. High‑fidelity grid simulators often simplify decentralized behavior~\cite{Iranpour2025UnifiedCoSim}, whereas agent‑based frameworks typically abstract electrical constraints via coarse aggregate bounds~\cite{Christensen2024MABSEV,Cong2025MABS}. Few shared‑charging studies enforce ramping or forecast‑friendliness as explicit, real‑time constraints, and the relationship between user‑centric (SoC) and grid‑centric indicators is rarely analyzed in a way that keeps their objectives distinct~\cite{Obinata2024USIM}. Addressing these gaps calls for models that integrate realistic agent‑level behavior with explicit power‑fluctuation limits~\cite{Iranpour2025UnifiedCoSim}, produce forecast‑friendly aggregate demand~\cite{Cong2025MABS}, and sustain desirable SoC outcomes across heterogeneous user groups~\cite{Obinata2024USIM}.
\section{\uppercase{System Model}}
\label{sec:system_model}

\begin{figure}[ht]
\centering
\includegraphics[width=1\linewidth]{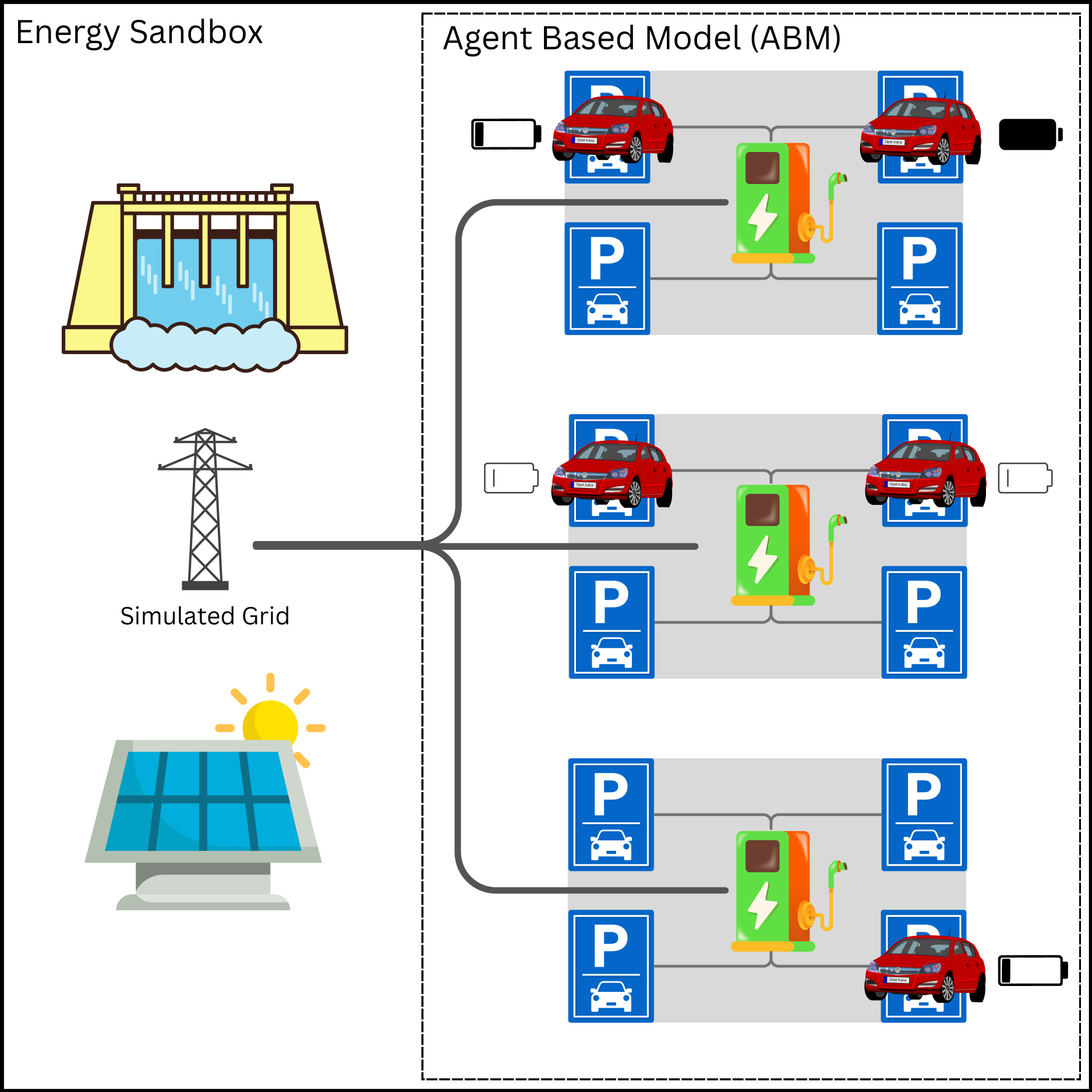}
\caption{ABM of the proposed EV charging framework, illustrating the interactions between EVs and CCs. The ABM is embedded within an ES, enabling the evaluation of different load scenarios under various operational constraints.}
\label{fig:system_model}
\end{figure}

This section presents the comprehensive ABM developed to simulate the dynamics of an EV charging system. The model captures both the temporal behavior of EVs, such as arrival times, parking durations, and individual charging demands, and the operational characteristics of CCs, including maximum power ratings and port availability. To enable grid-aware evaluations, the ABM operates within a basic Energy Sandbox (ES) that provides price signals and grid consumption limits. The entire framework is implemented in Python using the SimPy framework \cite{10.7717/peerj-cs.103}, enabling flexible and scalable process-oriented simulations across varying system scales. Figure~\ref{fig:system_model} illustrates the main model components and their interconnections.

\subsection*{EV Dynamics: Arrival, Parking, and Charging Eligibility}

To describe the behavioral processes governing individual EVs, including how they arrive, wait, park, and become eligible for charging within the facility, we define a finite set of EVs as
\[
\mathcal{V} = \{EV_1, EV_2, \dots, EV_{|\mathcal{V}|}\},
\]
where $|\mathcal{V}|$ denotes the total number of vehicles considered within the charging facility during the operating interval
\[
\mathcal{T} = [t_{\mathrm{s}},\, t_{\mathrm{e}}].
\]

Each $EV_i \in \mathcal{V}$ is associated with an entrance delay $T_{\mathrm{e}}^{i} \ge 0$. 
The actual arrival time of vehicle $i$ is therefore
\[
t_{\mathrm{arr}}^{i} = t_{\mathrm{s}} + T_{\mathrm{e}}^{i}.
\]

The set $\{T_{\mathrm{e}}^{i}\}_{i=1}^{|\mathcal{V}|}$ is parameterized according to the 
considered usage scenario, enabling the representation of different arrival 
patterns. For instance, under an average employee usage scenario, the majority 
of vehicles arrive during a morning time window (e.g., between 06:00 and 09:00). 
Such behavior is captured by sampling $T_{\mathrm{e}}^{i}$ accordingly within 
that interval.

Upon arrival at time $t_{\mathrm{arr}}^{i}$, $EV_i$ immediately requests a 
parking space equipped with an available CC port. 
Let $T_{\mathrm{w}}^{i}$ denote the maximum waiting time tolerated by vehicle $i$.

If a CC port becomes available within the waiting interval
\[
[t_{\mathrm{arr}}^{i},\, t_{\mathrm{arr}}^{i} + T_{\mathrm{w}}^{i}],
\]
the vehicle is admitted and assigned to that port. 
The time of successful assignment and physical connection is denoted by 
$t_{\mathrm{connect}}^{i}$.

At time $t_{\mathrm{connect}}^{i}$, an arrival event signal is transmitted to the 
corresponding CC, indicating that the port is occupied and that 
the vehicle is connected. If no port becomes available within the waiting 
interval, the vehicle leaves the facility without receiving service.

Upon successful admission, the vehicle occupies the assigned parking space 
for a parking duration $T_{\mathrm{p}}^{i}$. The departure time is therefore
\[
t_{\mathrm{dep}}^{i} = t_{\mathrm{connect}}^{i} + T_{\mathrm{p}}^{i}.
\]

The parking duration $T_{\mathrm{p}}^{i}$ is modeled as a stochastic parameter 
reflecting user behavior under the considered usage scenario. For example, 
under an average employee usage scenario, parking times typically correspond 
to a full working day. Such variability is captured 
by sampling $T_{\mathrm{p}}^{i}$ in accordance with the scenario assumptions.

Charging becomes eligible immediately upon successful parking and connection 
to the CC port, i.e., at time $t_{\mathrm{connect}}^{i}$. The vehicle remains 
connected and available for charging until its departure at $t_{\mathrm{dep}}^{i}$, 
subject to the model-defined charging control mechanism.

\subsection{Charging Columns and Energy Sandbox Interactions}

Here, we outline the operational logic of the charging infrastructure and its coordination with the Energy Sandbox, detailing how charging decisions, power allocation, and constraints are managed at the system level.

Let
\[
\mathcal{C} = \{CC_1, CC_2, \dots, CC_{|\mathcal{C}|}\}
\]
denote the set of CCs available in the facility. 

Each Charging Column $CC_j \in \mathcal{C}$ is characterized by a fixed number 
of charging ports and a maximum power rating. 
The number of ports per column is denoted by $N_{\mathrm{port}}$, and is set to
\[
N_{\mathrm{port}} = 4,
\]
implying that at most four vehicles may be physically connected to a given 
Charging Column at any time. 

The maximum deliverable charging power of $CC_j$ is denoted by $P_{j}^{\max}$, 
which defines the upper bound on the instantaneous charging power supplied 
to the actively selected vehicle, i.e.,
\[
0 < P_{j}(t) \le P_{j}^{\max}.
\]

Two types of Charging Columns are considered in this study. 
Slow Charging Columns (SCCs) operate with
\[
P_{j}^{\max} = 11 \ \text{kW},
\]
whereas Fast Charging Columns (FCCs) operate with
\[
P_{j}^{\max} = 48 \ \text{kW}.
\]
Each $CC_j$ is assigned to one of these two categories.

Although multiple vehicles may be connected simultaneously, 
only one vehicle can be actively charged at any given time. 
This operational constraint reflects the internal power allocation 
design assumed for the Charging Column.

For each Charging Column $CC_j$, the set of vehicles connected at time $t$ is defined as
\[
\mathcal{V}_j(t) 
= 
\{ EV_i \in \mathcal{V} \mid EV_i \text{ is connected to } CC_j \text{ at time } t \}.
\]

At each decision epoch, $CC_j$ selects one vehicle from $\mathcal{V}_j(t)$ 
to receive charging power according to the applied charging strategy.

Under the First-Come--First-Served (FCFS) strategy, vehicles are prioritized 
according to their connection times $t_{\mathrm{connect}}^{i}$. 
The selected vehicle at time $t$ is given by
\[
EV_{i^{*}}(t)
=
\arg\min_{EV_i \in \mathcal{V}_j(t)} t_{\mathrm{connect}}^{i}.
\]
The selected vehicle is continuously charged until its charging demand 
is fully satisfied, a leave request is issued, or its parking duration expires. 
Upon completion or disconnection, the next vehicle in the ordered set 
is selected according to the same rule.

Under the Time-Sharing strategy (SHRD), all vehicles in $\mathcal{V}_j(t)$ 
share access to the charging resource in a cyclic manner. 
At each electricity price interval of duration $\Delta t_{\mathrm{price}}$, 
one vehicle is selected in a round-robin fashion to receive charging power 
for the duration of that interval. Vehicles whose charging demand is fully 
satisfied or which disconnect are removed from the cyclic selection process. 
The procedure continues over the remaining connected vehicles 
until $\mathcal{V}_j(t)$ becomes empty.

Whenever $CC_j$ switches the active charging process from one vehicle 
to another, a handshake procedure is required between the Charging Column 
and the newly selected vehicle. This handshake introduces a switching delay 
denoted by $T_{\mathrm{hs}}$. In this study, the handshake duration is set to 
$T_{\mathrm{hs}} = 32$ seconds. Charging power delivery begins only after 
successful completion of the handshake. No additional handshake is required 
if the same vehicle continues charging without interruption.

At any time $t$ satisfying
\[
t_{\mathrm{connect}}^{i} \le t \le t_{\mathrm{dep}}^{i},
\]
a connected vehicle $EV_i$ may issue a leave request. 
Let $t_{\mathrm{leave}}^{i}$ denote the time of such a request. 
Upon receiving this request, the Charging Column immediately terminates 
the charging process of $EV_i$. The delivered energy is finalized, 
the corresponding charging cost is calculated based on accumulated energy 
and prevailing price signals, and the settlement information is communicated 
to the vehicle. The port is subsequently released and becomes available 
for future allocation.

Let the Energy Sandbox (ES) be defined as a facility-level resource that bounds the simultaneously deliverable charging power through a shared capacity
\[
P_{\mathrm{ES}}^{\max} \;>\; 0 .
\]
At any time $t$, let $p_j(t)\ge 0$ denote the power allocated by the ES to Charging Column $CC_j$, and define the aggregate allocated power as
\[
P_{\mathrm{alloc}}(t) \;=\; \sum_{CC_j \in \mathcal{C}} p_j(t),
\]
subject to the capacity constraint
\[
0 \;\le\; P_{\mathrm{alloc}}(t) \;\le\; P_{\mathrm{ES}}^{\max}, 
\qquad \forall t \in \mathcal{T}.
\]

Whenever $CC_j$ intends to actively charge a selected vehicle $EV_i$, a power request is issued to the ES. The requested power is denoted by $r_{j,i}(t)$ and is determined by the infrastructure and vehicle power limits. Let $P_{i}^{\max}$ denote the maximum charging power accepted by $EV_i$ (e.g., due to its onboard charger), and let $P_{j}^{\max}$ denote the maximum power rating of $CC_j$. The requested power is defined as
\[
r_{j,i}(t) \;=\; \min\!\big(P_{j}^{\max},\, P_{i}^{\max}\big).
\]
If time-varying vehicle acceptance limits are later considered, $P_{i}^{\max}$ can be interpreted as $P_{i}^{\max}(t)$ without changing the structure of the model.

A request issued at time $t_{\mathrm{req}}$ is granted if sufficient ES capacity is available, i.e., if
\[
P_{\mathrm{alloc}}(t_{\mathrm{req}}^{-}) + r_{j,i}(t_{\mathrm{req}}) \;\le\; P_{\mathrm{ES}}^{\max}.
\]
Upon grant, the allocation is updated by
\begin{align*}
p_j(t_{\mathrm{req}}) 
&= r_{j,i}(t_{\mathrm{req}}), \\
P_{\mathrm{alloc}}(t_{\mathrm{req}}) 
&= P_{\mathrm{alloc}}(t_{\mathrm{req}}^{-}) + p_j(t_{\mathrm{req}}).
\end{align*}
and charging may begin (subject to column-level constraints).

If insufficient power is available at $t_{\mathrm{req}}$, the request is queued and remains pending until it becomes feasible under the capacity constraint. Pending requests are served according to a First-Come--First-Served (FCFS) rule. Let
\[
\mathcal{Q}(t) = \big( (CC_{j_1}, EV_{i_1}, t_{\mathrm{req}}^{(1)}), (CC_{j_2}, EV_{i_2}, t_{\mathrm{req}}^{(2)}), \dots \big)
\]
denote the FCFS-ordered queue of outstanding ES power requests at time $t$. When ES capacity becomes available (e.g., due to a release), queued requests are granted in increasing order of their request times, as long as the capacity constraint remains satisfied.

During the waiting period, the connected vehicle may disconnect before the ES request is granted. Let $t_{\mathrm{leave}}^{i}$ denote the leave time of vehicle $EV_i$. If $CC_j$ is waiting for ES power to serve $EV_i$ and
\[
t_{\mathrm{leave}}^{i} < t_{\mathrm{grant}}^{j,i},
\]
where $t_{\mathrm{grant}}^{j,i}$ denotes the ES grant time associated with request $r_{j,i}(t)$, then the charging attempt is considered unsuccessful and the corresponding pending ES request is cancelled.

When the active charging episode at $CC_j$ ends (e.g., the vehicle disconnects), the allocated ES power is released at time $t_{\mathrm{rel}}$. The deallocation is modeled as
\begin{align*}
P_{\mathrm{alloc}}(t_{\mathrm{rel}}) 
&= P_{\mathrm{alloc}}(t_{\mathrm{rel}}^{-}) - p_j(t_{\mathrm{rel}}^{-}), \\
p_j(t_{\mathrm{rel}}) 
&= 0.
\end{align*}
thereby making the released capacity available for subsequent queued requests.

\begin{figure}[ht]
\centering
\includegraphics[width=1\linewidth]{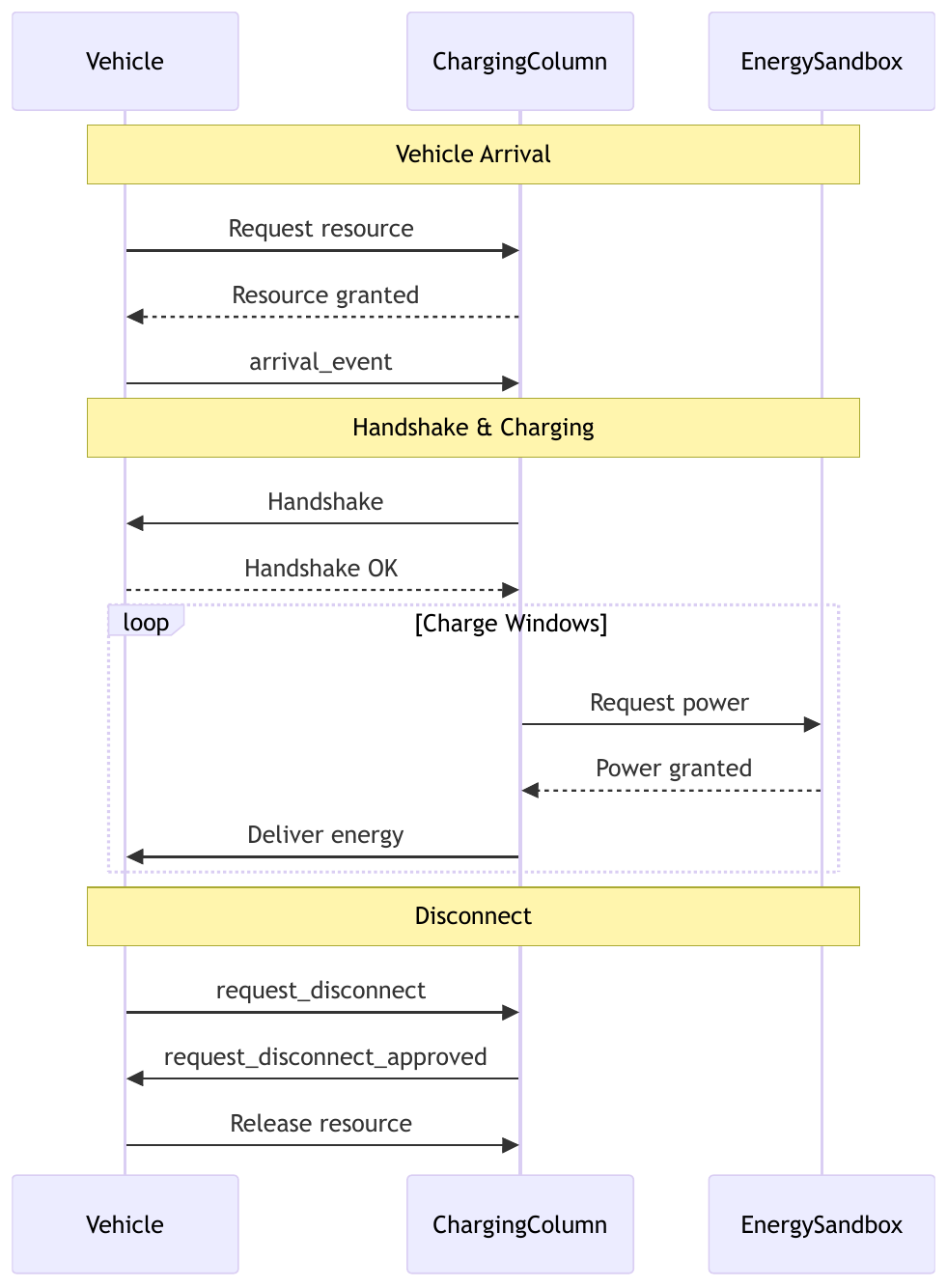}
\caption{Sequence diagram illustrating the interaction flow among an Electric Vehicle (EV), Charging Column (CC), and Energy Sandbox (ES) agents.}
\label{fig:interaction_flow}
\end{figure}

The interaction sequence among an EV, a Charging Column, and the Energy Sandbox is illustrated in Fig.~\ref{fig:interaction_flow}.

Neither of the charging strategies considered in the present study, namely FCFS and SHRD, incorporates time-varying photovoltaic (PV) availability or energy price information into its decision-making logic. Their charging behavior is determined solely by the respective service rule, without adaptation to external energy conditions. To maintain the completeness and extensibility of the proposed framework, the Energy Sandbox is designed to provide such exogenous information to the simulator, including energy prices and available PV power, thereby supporting the systematic investigation of more advanced charging approaches, such as self-organizing or learning-based methods. The corresponding time-series data must be supplied to the simulation environment prior to execution in the form of input CSV files.

\section{\uppercase{System Analysis}}
\label{sec:system_analysis}
This section reports the simulation results of the presented ABM for the EV charging system. The ABM is implemented in Python using the SimPy discrete-event simulation framework, and the complete source code is publicly available in a dedicated GitHub repository \cite{SharedCharging_ABM_repo}.
%\cite{youssefi2026}. 
The objective is to evaluate the operational behavior of the proposed framework under a practically relevant and well-defined usage scenario, namely an average employee workplace charging use case. All analyses are conducted within this scenario, which specifies representative assumptions on arrival times, parking duration, and charging demand, thereby defining the boundary conditions of the system. The results demonstrate the flexibility and scalability of the ABM by systematically varying charging column types, scheduling strategies, and system size. The impact of these configurations is examined with respect to the rate of energy delivery to EVs, the utilization characteristics of charging column circuits, and the aggregated behavior at the Energy Sandbox level, reflecting system-level power constraints and load dynamics.

The proposed ABM supports a broader set of configurable features, including heterogeneous charging voltages, dynamic energy pricing schemes, and extended operational policies. These aspects are not explored in the present experiments to maintain a focused analysis, but are implemented and available in the corresponding GitHub repository.
% ---- Use Case ----
\subsection*{Use case}
The experimental evaluation is conducted under an average employee usage scenario representing typical workplace charging conditions. 
Vehicle arrival times are restricted to a morning interval
\[
t_{\mathrm{arr}}^{i} \in [06{:}00,\;08{:}00],
\]
following a one-way commuting distance
\[
d_i = 26\ \text{km}.
\]

The energy demand upon arrival is determined by the commute-related consumption and satisfies
\[
E_i^{\mathrm{req}} \in [9,\;10]\ \text{kWh},
\]
corresponding to the energy deficit induced prior to connection.

The parking duration is defined within the working-day interval
\[
T_{\mathrm{p}}^{i} \in [8,\;9]\ \text{h},
\]
with departure time
\[
t_{\mathrm{dep}}^{i} = t_{\mathrm{arr}}^{i} + T_{\mathrm{p}}^{i}.
\]

Within this usage scenario, sufficient parking availability is assumed under nominal operating conditions. Nevertheless, the admission, waiting, and energy allocation mechanisms remain active in the model to preserve structural consistency with alternative usage scenarios considered in subsequent analyses.

While this study focuses on a representative workplace charging scenario, the proposed framework is not restricted to this context. The modular structure of EV arrival processes, parking duration distributions, and Energy Sandbox constraints enables straightforward extension to alternative usage patterns, including residential overnight charging, retail parking with high turnover, or grid-constrained feeder-level studies. Future work will systematically evaluate these scenario classes to analyze infrastructure suitability under differing temporal demand structures.

% ---- Experiments ----
\subsection*{Experimental setup}
A total of 12 experiments are defined to systematically analyze the behavior of the proposed ABM under different charging infrastructure configurations, scheduling strategies, and system scales. In particular, the number of EVs is varied to represent increasing load levels, while the infrastructure is configured either with FCC or SCC. In addition, two charging scheduling strategies are considered, namely FCFS and the proposed SHRD strategy. In the present study, these strategies are intentionally used as baseline rule-based charging policies to validate the framework and to establish reference system behavior for future comparisons with more advanced coordination approaches. The complete set of experiment configurations is summarized in Table~\ref{tab:exp_setups}. In all experiments, the maximum power of the Energy Sandbox, $P^{\max}_{\mathrm{ES}}$, is fixed to $1\,\mathrm{MW}$, deliberately exceeding the aggregate charging demand of the considered setups. Consequently, the grid constraint remains non-binding, allowing the analysis to isolate the effects of infrastructure configuration and charging scheduling strategies. The Energy Sandbox mechanism becomes critical under grid-constrained scenarios, which are part of ongoing work and future extensions of the presented framework.

\begin{table}[ht]
\centering
\begin{tabular}{c c c c c}
\hline
Exp. ID & EVs & FCC & SCC & Strategy \\
\hline\hline
1  & 30  & 30 & 0  & FCFS \\
2  & 30  & 30 & 0  & SHRD \\
3  & 30  & 0  & 30 & FCFS \\
4  & 30  & 0  & 30 & SHRD \\
\hline
5  & 60  & 30 & 0  & FCFS \\
6  & 60  & 30 & 0  & SHRD \\
7  & 60  & 0  & 30 & FCFS \\
8  & 60  & 0  & 30 & SHRD \\
\hline
9  & 120 & 30 & 0  & FCFS \\
10 & 120 & 30 & 0  & SHRD \\
11 & 120 & 0  & 30 & FCFS \\
12 & 120 & 0  & 30 & SHRD \\
\hline
\end{tabular}
\caption{Experiment setups for the simulation study. FCC: Fast Charging Columns; SCC: Slow Charging Columns; EVs: Number of Electric Vehicles; Strategy: Charging scheduling strategy.}
\label{tab:exp_setups}
\end{table}

All experiments listed in Table~\ref{tab:exp_setups} are repeated 30 times with independent random seeds to account for stochastic effects and ensure statistically robust results.

\subsection*{Simulation results}

To evaluate the energy delivery performance at the EV level, a unified Key Performance Indicator (KPI) is introduced. 
Since the individual required charging energy $E^{\mathrm{req}}_i$ varies across $\mathrm{EV}_i$ and simulation instances, a direct comparison of absolute charging completion times or total received energy is not suitable for aggregated analysis. 
Therefore, a fixed reference energy threshold is introduced to ensure a consistent and directly comparable performance metric across different system configurations.

For the considered average employee usage scenario, the mean commuting energy demand is approximately
\begin{equation*}
E^{\star} = 9.36\,\mathrm{kWh}.
\end{equation*}
This value follows directly from the scenario configuration and represents a practically satisfactory charging amount within the modeled context. It is emphasized that $E^{\star}$ is scenario-dependent and does not constitute a universal constant.

Based on this reference, the KPI $\mathrm{TTR}_{9.36}$ (Time-To-Receive $9.36\,\mathrm{kWh}$) for $\mathrm{EV}_i$ is defined as the smallest time $\Delta t_i$ after arrival such that
\begin{equation*}
\int_{t_{\mathrm{arr}}^i}^{t_{\mathrm{arr}}^i + \Delta t^i} 
P_i(\tau)\, d\tau 
= E^{\star},
\end{equation*}
where $t^{\mathrm{arr}}_i$ denotes the arrival time of $\mathrm{EV}_i$ and $P_i(t)$ is the instantaneous charging power allocated to $\mathrm{EV}_i$. 
Thus,
\begin{equation*}
\mathrm{TTR}_{9.36}^{i} = \Delta t^i.
\end{equation*}

\begin{figure}[ht]
\centering
\includegraphics[width=1\linewidth]{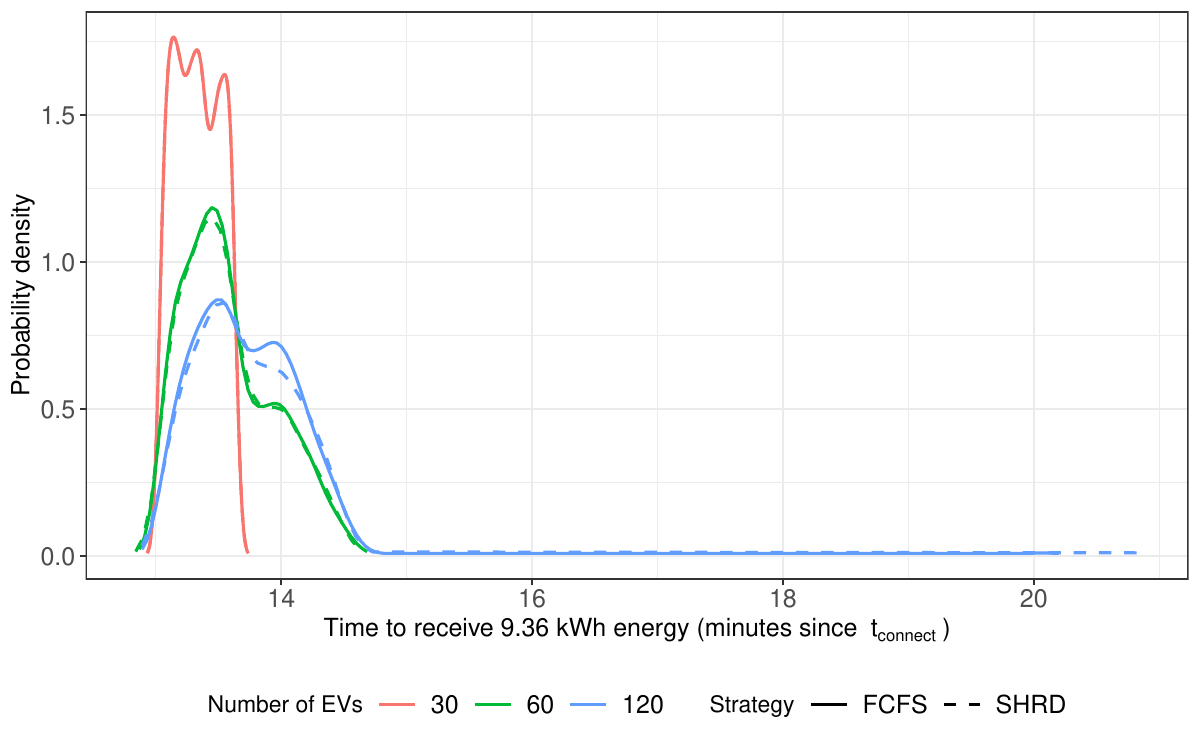}
\caption{Probability density of $\mathrm{TTR}_{9.36}$ for FCC configurations (Exp.~IDs~1, 2, 5, 6, 9, and~10), comparing different EV counts and charging strategies.}
\label{fig:ttr936_fcc_pdf}
\end{figure}

The distribution of $\mathrm{TTR}_{9.36}$ across all EVs provides an interpretable measure of the effective energy delivery dynamics of the system under different infrastructure configurations and charging strategies. Figure~\ref{fig:ttr936_fcc_pdf} shows the probability density of $\mathrm{TTR}_{9.36}$ for FCC configurations (Exp.~IDs~1, 2, 5, 6, 9, and~10). As shown in the figure, for the case of 30 EVs, both charging strategies exhibit nearly identical distributions, indicating that under low system load no noticeable performance difference emerges. 
In this regime, each charging column can serve its assigned EVs without frequent reallocation, resulting in uninterrupted power delivery. 
With increasing EV counts, the distributions slightly shift toward larger values of $\mathrm{TTR}_{9.36}$, reflecting longer delivery times. 
This effect is attributable to intensified contention at the charging columns, where power must be sequentially reassigned among multiple connected EVs, leading to additional switching-induced delays (i.e, handshakes).

Each density represented in Figure~\ref{fig:ttr936_fcc_pdf} highlights the most typical delivery times and the dispersion of charging performance across the EV population and charging scheduling strategy. As observed in the figure, the distribution for 30 EVs is relatively concentrated, while it broadens for 60 and 120 EVs, indicating increased variability in energy delivery times under higher system load.

\begin{figure}[ht]
\centering
\includegraphics[width=1\linewidth]{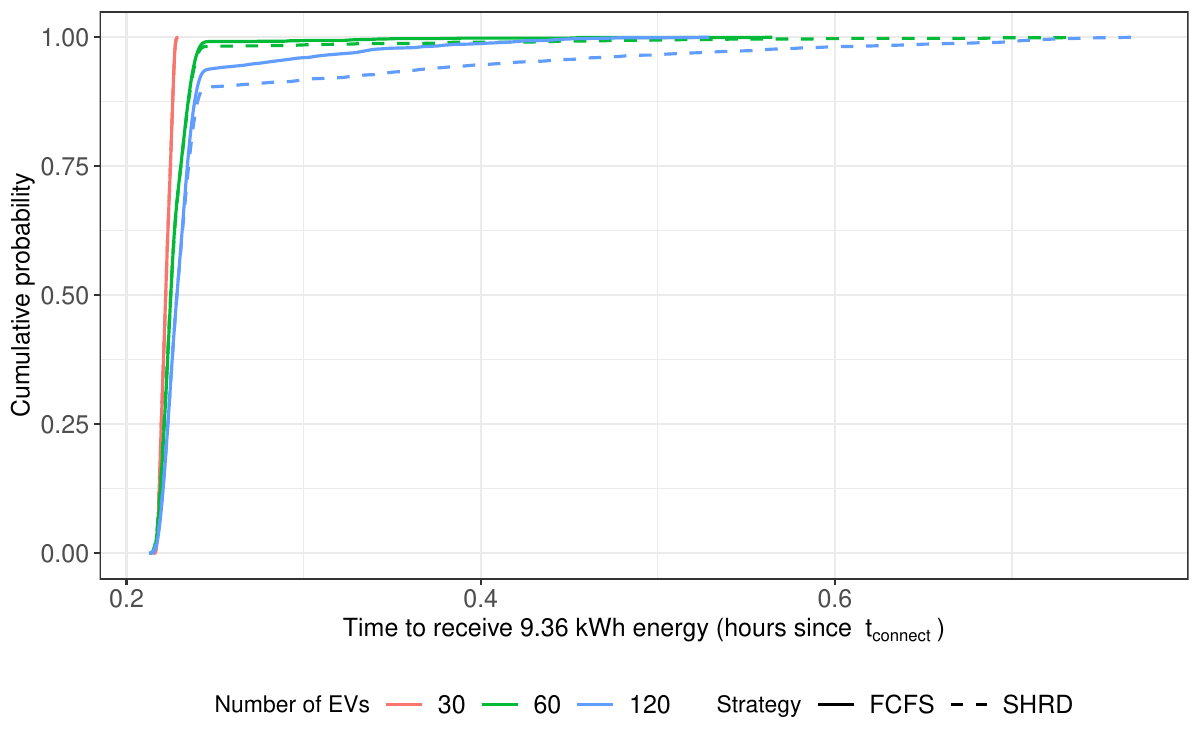}
\caption{Cumulative distribution of $\mathrm{TTR}_{9.36}$ for FCC configurations (Exp.~IDs~1, 2, 5, 6, 9, and~10), comparing different EV counts and charging strategies.}
\label{fig:ttr936_fcc_cdf}
\end{figure}

The same results are presented in cumulative form in Figure~\ref{fig:ttr936_fcc_cdf}, which shows the Cumulative Distribution Function (CDF) of $\mathrm{TTR}_{9.36}$. 
The CDF represents the probability that an $\mathrm{EV}_i$ receives the reference energy $E^{\star}$ within a given time $t$, formally defined as
\begin{equation*}
F_{\mathrm{TTR}_{9.36}}(t) 
= \mathbb{P}\!\left( \mathrm{TTR}_{9.36} \le t \right),
\end{equation*}
providing a direct and interpretable measure of system responsiveness under the considered configurations.
% ---- FCC-only TTR936 ----

\begin{figure}[ht]
\centering
\includegraphics[width=1\linewidth]{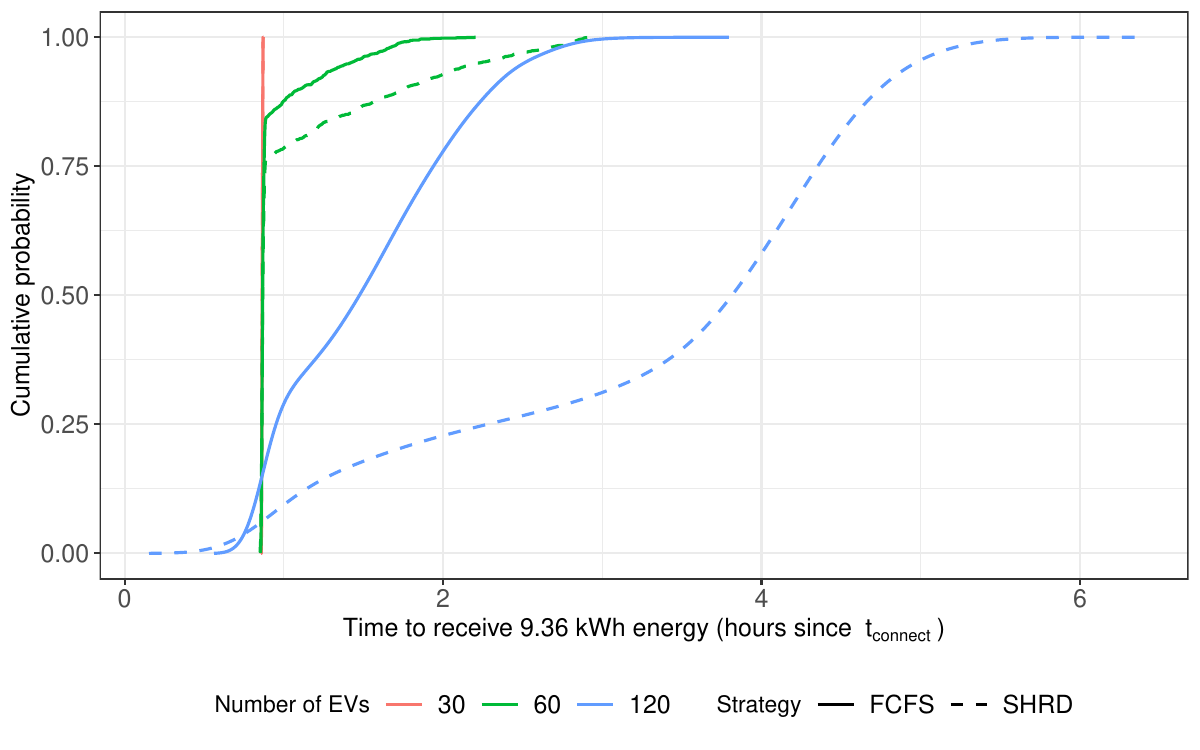}
\caption{Cumulative probability (CDF) of $\mathrm{TTR}_{9.36}$ for SCC configurations across EV counts and strategies.}
\label{fig:ttr936_scc_cdf}
\end{figure}

Figure~\ref{fig:ttr936_scc_cdf} presents the cumulative distribution of $\mathrm{TTR}_{9.36}$ for SCC configurations (Exp.~IDs~3, 4, 7, 8, 11, and~12) across different EV counts and charging strategies. 
Compared to the FCC case, the corresponding CDFs exhibit a noticeably smaller slope, reflecting the lower nominal charging power of slow charging columns. 
Since $\mathrm{TTR}_{9.36}^{(i)}$ is defined as the time required to accumulate the fixed energy $E^{\star}$, a reduced instantaneous charging power directly results in larger delivery times and therefore a flatter cumulative curve.

% ---- SCC-only TTR936 ----

A comparison of Figures~\ref{fig:ttr936_fcc_cdf} and~\ref{fig:ttr936_scc_cdf} reveals two notable effects. First, the degradation with increasing EV count is more pronounced than in the FCC case, indicating stronger sensitivity to load under limited power throughput, and thus benefiting scenarios where a grid power cap is enforced. Second, for high EV counts, the SHRD strategy results in a further rightward shift of the CDF. Under low per-window energy allocation, sequential reassignment of charging resources induces a larger number of switching events, increasing interruption overhead and prolonging the time required to reach $E^{\star}$. Consequently, strategies involving frequent reallocation become increasingly disadvantageous in low-power infrastructure settings for limited average parking time scenarios.

\begin{figure}[ht]
\centering
\includegraphics[width=1\linewidth]{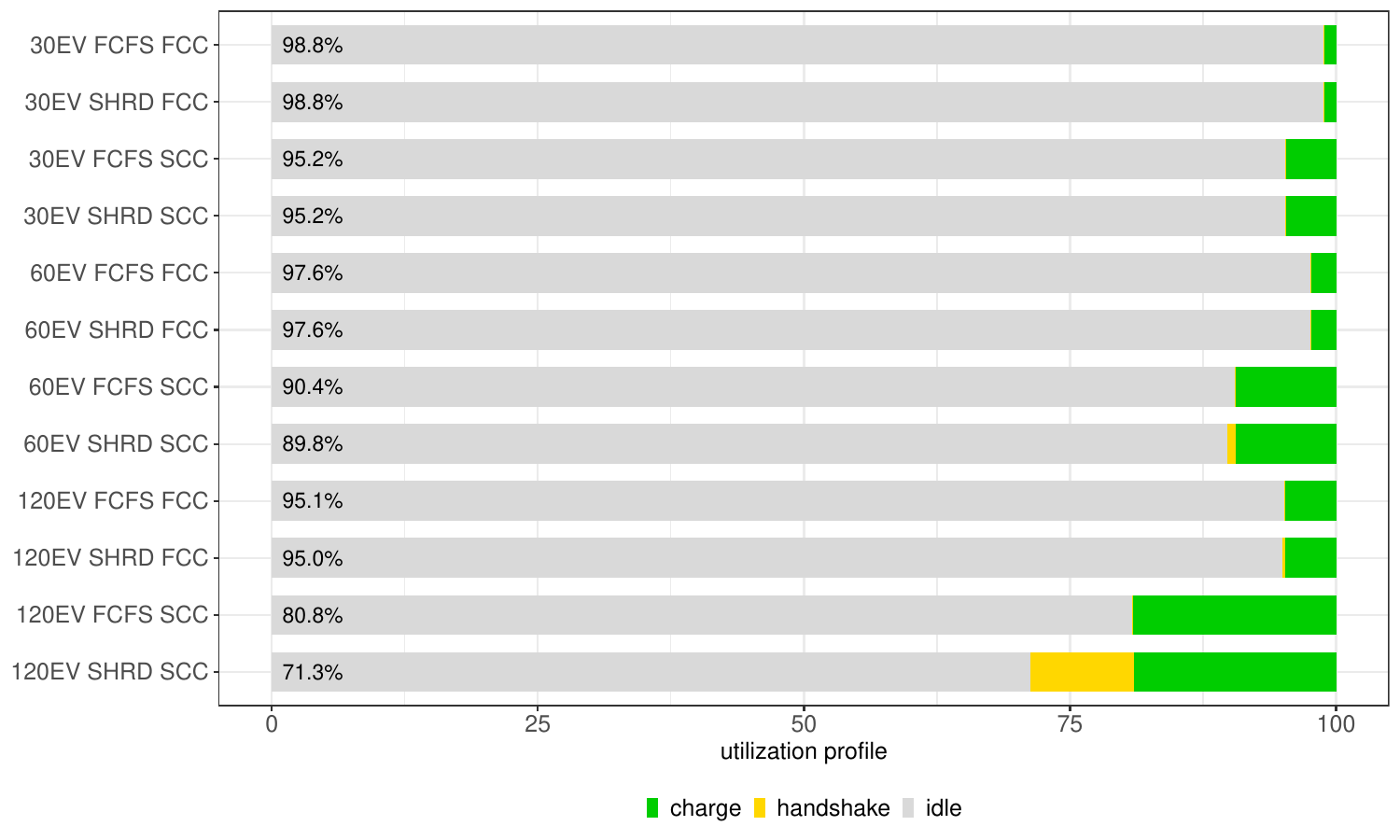}
\caption{Charging Column (CC) \textit{utilization profiles} aggregated per configuration.}
\label{fig:cc_state_bands}
\end{figure}

Figure~\ref{fig:cc_state_bands} presents the \textit{utilization profile} of charging columns for all experiment configurations listed in Table~\ref{tab:exp_setups}. 
For each experiment, a horizontal band illustrates the percentage of time spent in the three possible operational states: \textit{charge}, \textit{handshake}, and \textit{idle}. 
The horizontal axis represents the relative time share in percent, while the vertical axis enumerates the individual experiment configurations.

% ---- CC ----

\begin{figure*}[ht]
\centering
\includegraphics[width=0.8\linewidth]{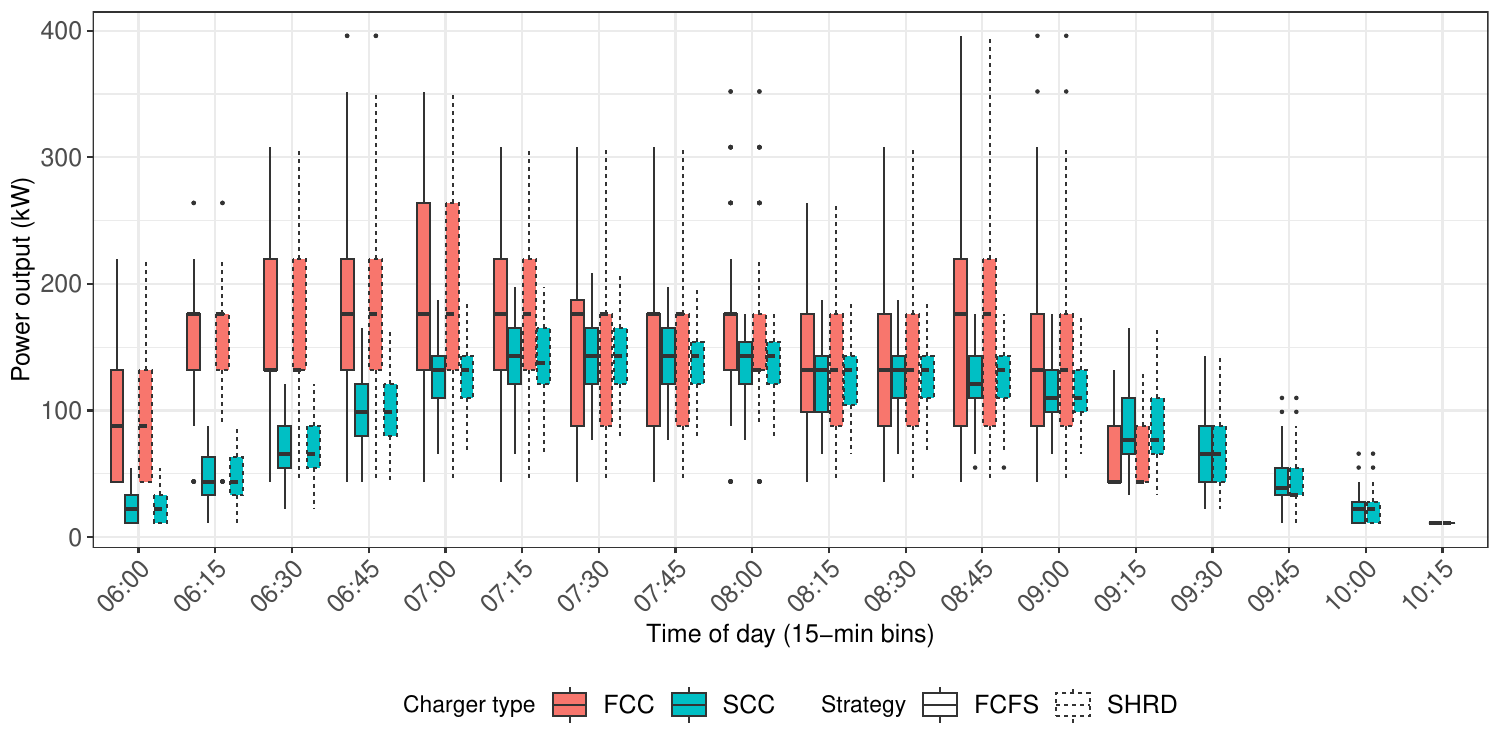}
\caption{Energy Sandbox power output distribution (15-min bins) for 30 EVs.}
\label{fig:pwr_box_30ev}
\end{figure*}

\begin{figure*}[ht]
\centering
\includegraphics[width=0.8\linewidth]{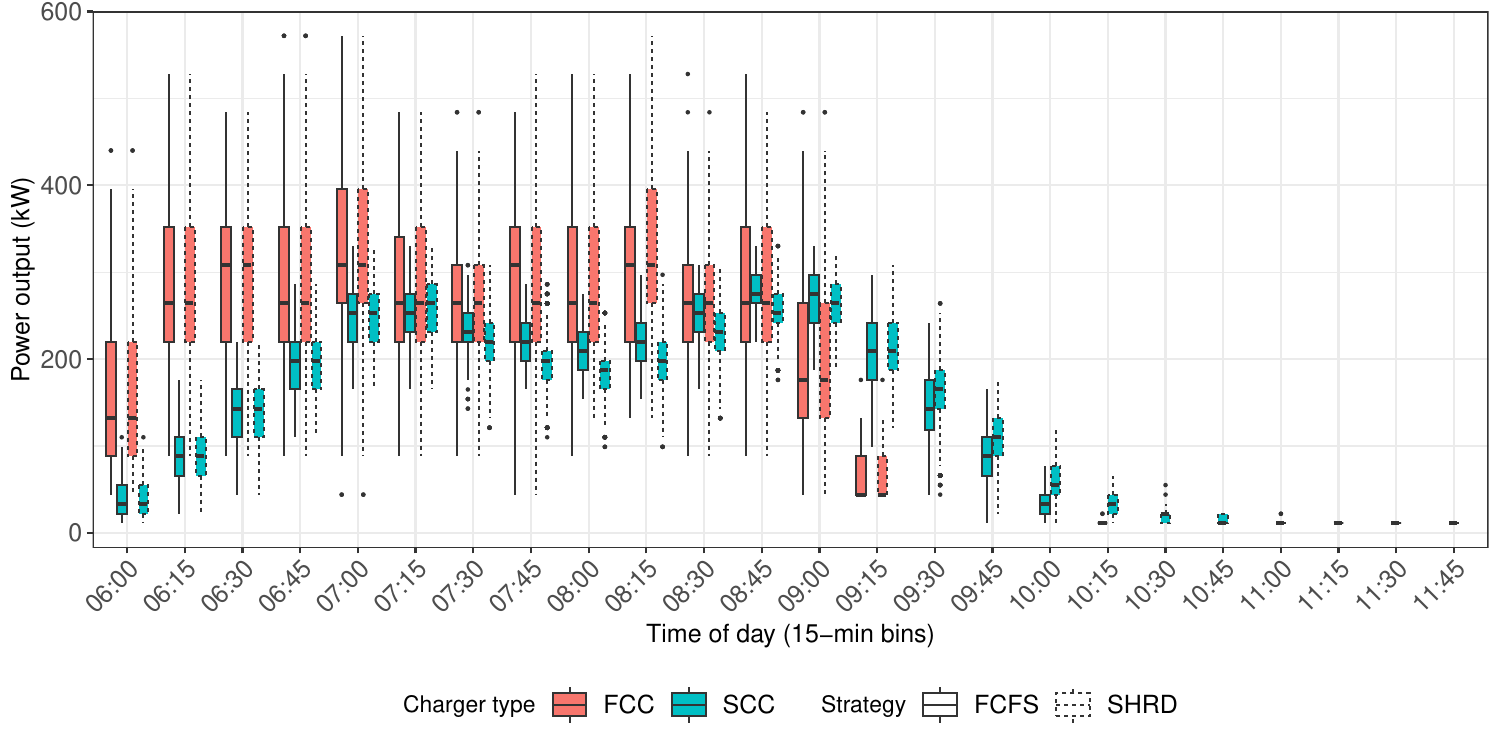}
\caption{Energy Sandbox power output distribution (15-min bins) for 60 EVs.}
\label{fig:pwr_box_60ev}
\end{figure*}

\begin{figure*}[ht]
\centering
\includegraphics[width=0.85\linewidth]{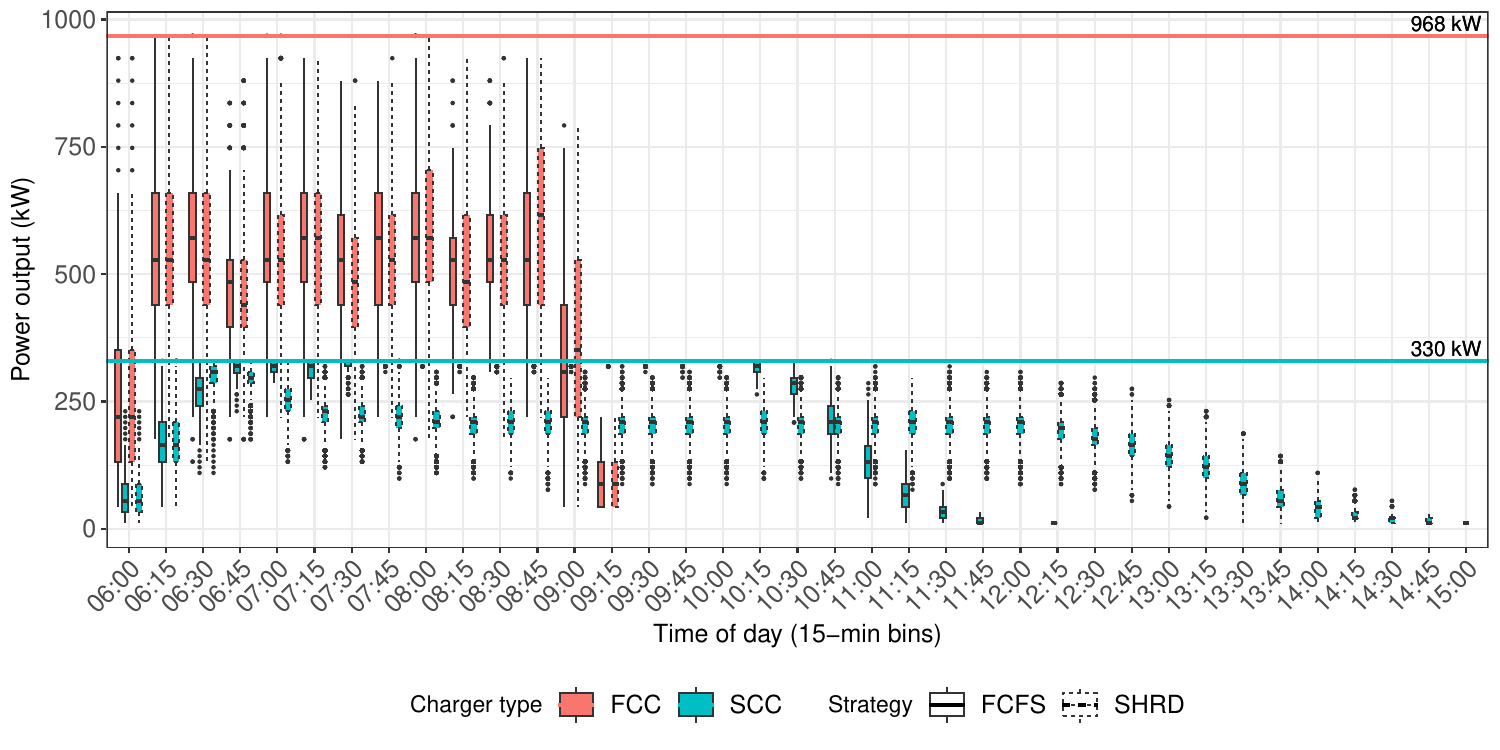}
\caption{Energy Sandbox power output distribution (15-min bins) for 120 EVs.}
\label{fig:pwr_box_120ev}
\end{figure*}

As shown in Figure~\ref{fig:cc_state_bands}, the idle fraction dominates the utilization profile across all configurations, in most cases approaching $90\%$. 
This behavior follows directly from the relation between the effective charging duration and the parking time in the considered use case. 
Let $T^{\mathrm{chg}}_i$ the time required to deliver its requested energy. 
Under the average employee scenario,
\begin{equation*}
t^i_{\mathrm{chg}} \ll T^i_{\mathrm{p}}, \quad \forall i.
\end{equation*}
where $t^i_{\mathrm{chg}}$ denotes the required charging time of vehicle $i$; this relation holds even for slow charging columns.
Consequently, charging columns remain idle for a substantial portion of the simulation horizon.
This observation is structurally tied to the scenario assumptions and does not indicate under-dimensioning of the infrastructure.

A second notable effect appears for the configuration with 120 EVs using slow charging columns under the SHRD strategy. 
In this case, the handshake fraction increases significantly compared to the corresponding FCFS configuration, while the pure charging fraction remains approximately unchanged. 
The difference is therefore primarily attributable to additional switching overhead. 
This behavior is consistent with the rightward shift of the corresponding SHRD curve observed in Figure~\ref{fig:ttr936_scc_cdf}, where increased handshake phases translate into delayed energy delivery.  In this context, the simulation outcomes demonstrate that the SHRD strategy yields operational benefits only under conditions in which the charging column is never transitioned into an \textit{idle} state, when the aggregate charging demand of all EVs parked throughout the day fully saturates the available charging capacity; in other words, when: 
\begin{equation*}
t_{\mathrm{chg}}^i \approx T_{\mathrm{P}}^i 
\end{equation*}
In this scenario, the target energy level $E^{\star}$ is not attained for all vehicles; however, the SHRD scheme guarantees that each vehicle in $\mathcal{V}_j(t)$ receives at least a minimum allocated energy amount if compared to FCFS where only the first cars which are served get the full $E^{\star}$. This reasoning may be extended in future work that systematically evaluates other scenario classes under varying temporal demand patterns, as noted; however, such analysis lies beyond the scope of this paper.

% ---- Power output ----
Figures~\ref{fig:pwr_box_30ev}--\ref{fig:pwr_box_120ev} present the Energy Sandbox power allocation results for system scales of 30, 60, and 120 EVs, respectively. 
Each figure shows boxplots of $P^{\mathrm{alloc}}_{ES}(t)$, evaluated over 15-minute intervals and aggregated across 30 independent simulation runs per configuration. 
The color coding distinguishes the charging column type (FCC vs.\ SCC), while the boxplot outline differentiates the charging scheduling strategy (FCFS vs.\ SHRD). 
These plots therefore summarize the temporal allocation behavior of the Energy Sandbox under varying infrastructure types, strategies, and load levels.

As shown in Figure~\ref{fig:pwr_box_30ev}, under a system scale of 30 EVs (i.e., low load), the Energy Sandbox allocation $P^{\mathrm{alloc}}_{ES}(t)$ exhibits nearly identical distributions across charger types and scheduling strategies, indicating comparable aggregate power demand at the facility level. 
Figures~\ref{fig:ttr936_fcc_cdf} and~\ref{fig:ttr936_scc_cdf} confirm that although SCC configurations yield larger $\mathrm{TTR}_{9.36}$ values than FCC configurations, the delivery times remain well within the available parking duration. 
This is consistent with the utilization profiles in Figure~\ref{fig:cc_state_bands}, where charging columns are predominantly idle, implying that charging demand does not saturate infrastructure capacity. 
Taken together, these observations indicate that the additional power capability of FCC infrastructure does not translate into a meaningful operational advantage under low-load conditions. Considering the typically higher installation and grid-integration requirements of fast charging systems, their deployment appears disproportionate to the performance gains observed within the analyzed workplace scenario.

Under moderate load (60 EVs), Figure~\ref{fig:pwr_box_60ev} shows a clearer separation between FCC and SCC configurations from a grid perspective. 
FCC deployments lead to higher average values of $P^{\mathrm{alloc}}_{ES}(t)$ and exhibit steeper ramp-up and ramp-down behavior, whereas SCC configurations operate at lower and smoother power levels. 
Despite these distinct power characteristics, for both configurations the effective charging completion times remain well within the available parking window, and Figure~\ref{fig:cc_state_bands} indicates that FCC infrastructure is even more idle (approximately 7\%) than its SCC counterpart at this load level. 
Moreover, the difference between FCFS and SHRD remains marginal at the Energy Sandbox level, suggesting that scheduling strategy has limited influence on aggregate grid behavior under moderate load. 
Taken together, the additional peak demand induced by FCC infrastructure does not translate into a proportional operational benefit, while incurring substantially higher investment cost.

Under high load (120 EVs), the divergence becomes more pronounced, as shown in Figure~\ref{fig:pwr_box_120ev}. 
FCC configurations generate substantially higher power peaks (on the order of $960\,\mathrm{kW}$), while SCC configurations remain near approximately one-third of that level (around $330\,\mathrm{kW}$). 
Although FCC infrastructure reduces individual charging times, the parking duration remains sufficient to accommodate slower charging without service degradation. 
Consequently, the higher power capability primarily manifests as amplified grid stress rather than improved functional performance. 
In addition, FCC deployments exhibit sharper power transients, indicating more aggressive load fluctuations. 
Thus, even under high load, the increased peak demand and volatility associated with FCC infrastructure are not accompanied by a decisive operational advantage within the considered usage scenario. 

In greater detail, the higher cost of FCC versus SCC stems not only from the charging column hardware but also from the substantially more demanding power distribution infrastructure, such as larger cable cross‑sections, enhanced cooling, and dedicated switchgear. Additionally, Medium Voltage (MV) grid connection may become technically or economically unviable once the site’s aggregate power demand surpasses defined capacity limits.
Given the substantially higher capital cost of fast charging columns, the results consistently support SCC infrastructure as the more suitable solution for this application context.

\begin{figure}[ht]
\centering
\includegraphics[width=1\linewidth]{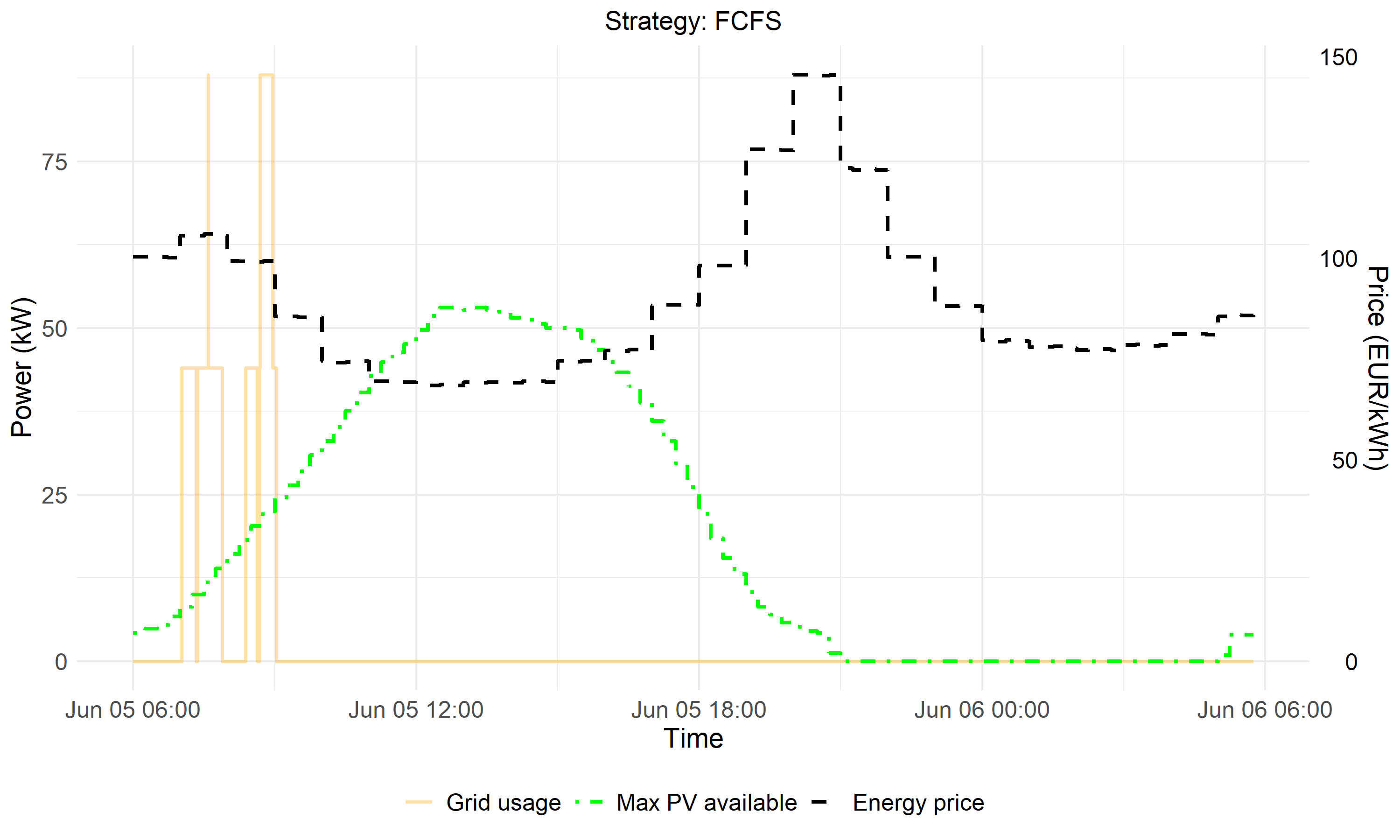}
\caption{Grid usage for the FCFS strategy, shown together with PV availability and energy price over time in the average employee scenario.}
\label{fig:GridPVaPrice}
\end{figure}

To further illustrate the interaction between charging demand and external energy conditions, Figure \ref{fig:GridPVaPrice} presents a representative FCFS run together with the corresponding time-varying PV availability and energy price signals. As a representative traditional charging strategy, FCFS does not explicitly incorporate either of these signals into its decision-making process. More generally, traditional charging strategies determine charging behavior primarily according to predefined service rules rather than renewable availability or price-favorable periods. In the considered workplace scenario, this leads to a temporal mismatch between charging demand and external energy conditions, since charging demand is concentrated early in the day whereas PV availability emerges later. This example therefore highlights the usefulness of the proposed framework as a foundation for subsequent studies on more advanced charging approaches that explicitly account for such external energy signals.

\section{\uppercase{Conclusions}}
\label{sec:conclusion}

This paper presented a comprehensive Agent-Based Model for the analysis of EV charging systems under a representative workplace usage scenario. 
The model captures the interaction between EV agents, Charging Columns, and the Energy Sandbox within a discrete-event simulation framework, enabling a structured evaluation of infrastructure configurations and charging scheduling strategies. 
Based on the defined KPI $\mathrm{TTR}_{9.36}$, utilization profiles, and system-level power allocation behavior, the results demonstrate that infrastructure dimensioning must be aligned with the actual usage context. 
In the considered average employee scenario, slow charging columns provide good performance even under increased load, while fast charging infrastructure primarily induces higher grid peaks and overall infrastructure costs without delivering proportional operational benefit.

Beyond the specific findings of this study, the presented ABM constitutes a highly configurable and extensible research framework. 
System scale can be adjusted in terms of the number of EVs, charging columns, and their technical parameters, allowing the investigation of diverse infrastructure layouts and load conditions. 
Charging scheduling strategies can be modified or replaced to evaluate alternative coordination mechanisms, including priority-based, decentralized, or learning-based approaches.

At the system level, the integrated Energy Sandbox enables the modeling of grid-side behavior, including explicit capacity limits, dynamic constraints, and external control signals. 
It further supports the incorporation of energy price signals, thereby enabling the assessment of price-aware charging decision strategies and economically driven coordination policies.

The implementation in the SimPy discrete-event framework ensures high computational efficiency, as system dynamics are processed event-driven rather than through fixed time-step evaluation, resulting in significantly improved scalability compared to conventional time-based simulation approaches. 
Moreover, stochastic arrival patterns, heterogeneous EV characteristics, and scenario-dependent behavioral rules can be incorporated without structural modifications to the model architecture.

Collectively, this flexibility enables systematic exploration of a wide range of operational, infrastructural, economic, and policy-oriented questions, extending well beyond the specific configuration analyzed in this work. 
Accordingly, the proposed ABM provides a highly configurable simulation environment for studying EV charging ecosystems under diverse technical and operational conditions and serves as a simulation foundation for subsequent studies on advanced charging coordination. Future work will investigate the integration of decentralized, self-organizing, and learning-based charging strategies within the proposed ABM under signal-aware and grid-constrained operation.

\section*{\uppercase{Acknowledgements}}
This work was performed in the course of project Shared Charging supported by FFG under contract number 926939. ChatGPT \cite{openai_chatgpt_2026} was used to assist with language editing, readability improvement, and phrasing refinement. All technical content, interpretations, and conclusions were developed and verified by the authors.

\bibliographystyle{apalike}
{\small
\bibliography{bibliography}}

% \section*{\uppercase{Appendix}}

% If any, the appendix should appear directly after the references without numbering, and not on a new page. To do so please use the following command: \textit{$\backslash$section*\{APPENDIX\}}

\end{document}